\documentclass[10pt,twocolumn,letterpaper]{article}

\usepackage{wacv}              

\usepackage{graphicx}
\usepackage{booktabs,pifont,makecell}
\newcommand{\cmark}{\ding{51}} \newcommand{\xmark}{\ding{55}}

\definecolor{wacvblue}{rgb}{0.21,0.49,0.74}
\usepackage[pagebackref,breaklinks,colorlinks,allcolors=wacvblue]{hyperref}

\def\wacvPaperID{1092} 
\def\confName{WACV}
\def\confYear{2027}

\title{VERA-Hand: A Verified Multidimensional Benchmark for 2D Hand Pose Estimation in Third-Person Dynamic Gesture Videos}

\author{Gibran Benitez-Garcia$^{1}$ and Hiroki Takahashi$^{1,2,3}$\\ [1ex]
$^{1}$Graduate School of Informatics and Engineering\\
$^{2}$Artificial Intelligence eXploration Research Center (AIX)\\
$^{3}$Meta-Networking Research Center (MEET)\\
The University of Electro-Communications, Tokyo, Japan\\
{\tt\small gibran@ieee.org, rocky@inf.uec.ac.jp}
}

\begin{document}
\maketitle
{\let\thefootnote\relax\footnotetext{\copyright~2027 IEEE. Personal use of this material is permitted. Permission from IEEE must be obtained for all other uses, in any current or future media, including reprinting/republishing this material for advertising or promotional purposes, creating new collective works, for resale or redistribution to servers or lists, or reuse of any copyrighted component of this work in other works. Accepted to WACV 2027.}}

\begin{abstract}
Hand landmarks serve as input for gesture recognition, sign language understanding, and touchless human--computer interaction. 
Yet hand pose estimators are typically evaluated with frame-wise localization metrics such as PCK or MPJPE, which are computed over matched predictions and overlook missed or spurious hands, handedness errors, and temporal instability.
To address these limitations, we present VERA-Hand, a benchmark that evaluates estimators using manually verified, frame-level 2D annotations for 10{,}690 hands across four complementary dimensions (detection, localization, identification, and temporal stability), as well as a joint Pose-F1 metric that requires both correct detection and accurate localization.
Evaluating six estimators spanning hand-specific, whole-body, reconstruction-based, and foundation-model paradigms, we find that no estimator dominates: rankings reorder sharply across dimensions.
VERA-Hand also exposes failures that localization-only and downstream evaluations may not reveal, including cross-hand overlap in over 85\% of frames for a recent foundation model.
A post-hoc cleanup case study demonstrates how the benchmark can validate corrective interventions, while stratified analyses show that robustness varies across motion-blur classes and hand regions.
We release the verified annotations, protocol, and code at \url{https://gibranbenitez.github.io/VERA_Hand/}.
\end{abstract}

\begin{figure}[t]
  \centering
  \includegraphics[width=0.98\linewidth]{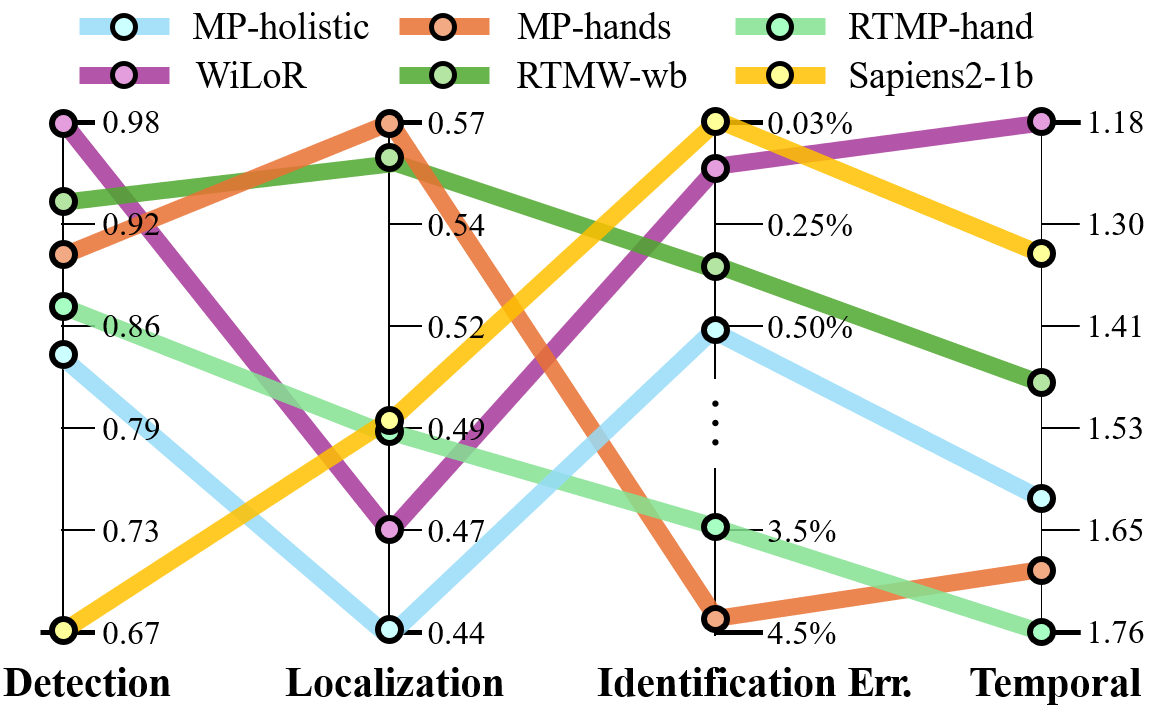}
  \caption{Ranking of the six evaluated estimators on VERA-Hand across four evaluation axes. Within each axis, estimators are ordered from best to worst, and scores are not directly comparable across axes (Detection and Localization $\uparrow$; Identification Error and Temporal Jitter $\downarrow$). No estimator ranks first across all four axes, and the ordering changes from one axis to another.}
  \label{fig:teaser}
\end{figure}

\begin{table*}[t]
  \centering
  \footnotesize
  \setlength{\tabcolsep}{4pt}
  \renewcommand{\arraystretch}{1.15}
  \resizebox{\textwidth}{!}{%
  \begin{tabular}{l c l l l l cccc l r}
    \toprule 
    & & & & & Human & \multicolumn{4}{c}{Annotation Support} & & \\ 
    \cmidrule(lr){7-10} 
    Dataset & Year & View & Content Type & GT Source 
      & Verified & Det & Loc & Id & Tmp & Evaluation & \#frames \\ 
    \midrule 
    FreiHAND~\cite{freihand}            & 2019 & Multi-view & Isolated images          & Multi-view (3D)      & Partial  &        & \cmark &        &        & Loc: PCK-AUC & 33K \\ 
    COCO-WholeBody~\cite{cocowholebody}      & 2020 & 3rd-person & Isolated images     & HRNet init (2D)       & All      &   \cmark  & \cmark & \cmark &        & Loc: AP/AR & 200K \\ 
    IPN Hand~\cite{ipnhand,ipnhandS} & 2020 & 3rd-person & Cont. gesture vids.   & MP-holistic (2D)      & Partial       &  & \cmark$^*$ &  & \cmark & HGR: Acc/Lev & \textemdash \\ 
    How2Sign~\cite{how2sign}            & 2021 & 3rd-person & Cont. SL videos  & OpenPose (2D)       & \xmark    &     & \cmark$^*$ &  & \cmark & SLT: Acc/BLEU & \textemdash \\ 
    ARCTIC~\cite{arctic}              & 2023 & Multi-view & HOI lab sequences       & Mocap (3D)           & \xmark            &        & \cmark & \cmark$^\dagger$ & \cmark & Loc: HOI-CDev & 2.1M \\ 
    AssemblyHands~\cite{assemblyhands}       & 2023 & Egocentric & Sampled videos      & Multi-view (3D)       & Partial  &        & \cmark &  \cmark$^\dagger$ &  \cmark & Loc: MPJPE & 3.0M \\ 
    iSign~\cite{isign}               & 2024 & 3rd-person & Cont. SL videos      & MP-holistic (2D)     & \xmark      &        & \cmark$^*$ &  & \cmark & SLT: Acc/BLEU & \textemdash \\ 
    HInt~\cite{hint}                 & 2024 & Exo \& Ego & Images+sequences & MMPose init (2D) & All & \cmark$^\dagger$ & \cmark & \cmark$^\dagger$ & \cmark$^\ddagger$ & Loc: PCK & 40.4K \\ 
    Ego-Exo4D (EgoPose)~\cite{egoexo4d} & 2024 & Ego \& Exo & Cont. videos           & Multi-view (3D)    & All      & \cmark$^\dagger$ & \cmark & \cmark$^\dagger$ & \cmark & Loc: PA-/MPJPE & 68K \\ 
    AnyHand~\cite{anyhand}             & 2026 & Synthetic  & Isolated images      & Synthetic (3D)   & \textemdash       & \cmark & \cmark & \cmark &        & Loc: PA-MPJPE & 6.6M \\ 
    Isharah~\cite{isharah}             & 2026 & 3rd-person & Cont. SL videos     & MP-holistic (2D)     & \xmark      &        & \cmark$^*$ &  & \cmark & CSLR: WER & \textemdash \\ 
    \midrule 
    \textbf{VERA-Hand} (built on~\cite{ipnhand}) & 2026 & 3rd-person & Cont. gesture vids. & MP-holistic init & All & \cmark & \cmark & \cmark & \cmark & \textbf{Det/Loc/Id/Tmp} & 10{,}768 \\ 
    \bottomrule
  \end{tabular}%
  }
  \caption{Comparison of hand pose datasets and annotation resources. 
  \emph{Evaluation} summarizes the original evaluation performed in each work, whereas
  \emph{Annotation Support} indicates which dimensions could be evaluated from released annotations (Det: detection, Loc: localization, Id: identification, Tmp: temporal stability), independently of the evaluation. 
  \emph{GT Source} identifies how the annotations were produced, and \emph{Human Verified} indicates the extent of human verification: All, Partial, or None (\xmark).
  \cmark$^*$ denotes support based on unverified predictions that may inherit model-specific errors. 
  \cmark$^\dagger$ denotes support tied to designated/target hand(s); full-frame all-hand exhaustiveness is not established. 
  \cmark$^\ddagger$ denotes temporal support only for samples with associated video sequences.}
  \label{tab:datasets}
\end{table*}

\section{Introduction}
\label{sec:intro}

Hand pose estimation is a core component of hand gesture recognition, sign language understanding, augmented and virtual reality, and touchless human--computer interaction~\cite{ipnhand,phoenix,how2sign,isharah}.
These applications rely on estimated hand landmarks as their input representation, and the quality of those keypoints has been shown to directly affect downstream performance~\cite{obrien-slt-eval,assemblyhands}.
Current estimators, however, span markedly different paradigms that are commonly used as alternatives: hand-specific models such as MediaPipe Hands (MP-hands)~\cite{mp-hands} and RTMPose-Hand (RTMP-hand)~\cite{rtmpose}; whole-body estimators that recover the hands jointly with the body and face, such as MediaPipe Holistic (MP-holistic)~\cite{mp-holistic, mediapipe} and RTMW whole-body (RTMW-wb)~\cite{rtmw}; reconstruction-based models that recover a parametric 3D hand representation, such as WiLoR~\cite{wilor}; and human-centric foundation models such as Sapiens2~\cite{sapiens2}.
They differ in how they detect, localize, and represent hands. Despite this heterogeneity, estimator selection is frequently treated as an implementation detail, while the way reliability is measured has not kept pace.

However, common evaluation practice remains centered on frame-wise localization metrics.
Hand-specific estimators are commonly assessed using Percentage of Correct Keypoints (PCK) and Mean Per Joint Position Error (MPJPE) \cite{freihand,interhand26m}, while whole-body models are typically evaluated with AP and AR based on Object Keypoint Similarity (OKS) \cite{cocowholebody}, mostly on isolated images, controlled sequences, or curated benchmarks.
Since localization is computed over predictions already matched to the ground truth (GT), these evaluations largely take hand detection as given and do not account for missed hands or spurious predictions.
They also provide limited insight into handedness identification and temporal stability, allowing estimators with markedly different failure patterns to receive similar aggregate scores.

This limitation is particularly important for continuous dynamic gestures, where hand scale varies, rapid motion produces blur, hands pass in front of the face, partially leave the image, and may be absent altogether.
Existing estimator comparisons in related settings rely on downstream task performance, controlled perturbations, or frame-level spatial evaluation rather than a unified verified reference \cite{obrien-slt-eval,assemblyhands,wang-metamorphic,hint}.
Meanwhile, gesture and sign language corpora increasingly release hand poses generated by off-the-shelf estimators, but without full human verification \cite{how2sign,isign,isharah}.
As summarized in Table~\ref{tab:datasets}, to the best of our knowledge,
no prior resource combines fully verified frame-level 2D annotations with a unified evaluation of detection, localization, identification, and temporal stability in a third-person continuous dynamic-gesture setting.

To address this gap, we introduce VERA-Hand, a benchmark for evaluating hand pose estimators on third-person, continuous, dynamic-gesture RGB videos.
Built on the IPN Hand dataset~\cite{ipnhand}, VERA-Hand provides manually verified, frame-level 2D hand pose annotations for 10{,}690 hands in 10{,}768 frames.
We evaluate six off-the-shelf estimators spanning four major paradigms across detection, localization, identification, and temporal stability.
Pose-F1 additionally evaluates detection and localization jointly, while stratified analyses examine performance across motion-blur classes and specific hand regions.
Our central finding is that estimator rankings change substantially across evaluation axes.
As Figure~\ref{fig:teaser} shows, WiLoR leads in detection but ranks second-last in localization, while MP-hands leads in localization despite a substantially higher identification error. No single estimator dominates across dimensions, demonstrating that localization alone provides an incomplete account of hand pose estimator reliability.

Beyond aggregate rankings, VERA-Hand exposes estimator-specific failure modes, most notably cross-hand overlap, in which a second prediction with the opposite handedness is overlaid on the only visible hand.
To demonstrate the benchmark's diagnostic utility, we include a targeted, training-free post-hoc cleanup case study.
Its spatial and temporal stages address cross-hand overlap, duplicate predictions, and detection gaps.
The intervention substantially improves the affected estimators and reorders the detection and Pose-F1 rankings, while producing only minor changes in pipelines that do not exhibit the targeted failures.

Our contributions are:
\begin{itemize}
    \item VERA-Hand, a manually verified benchmark with frame-level 2D hand pose annotations for 10{,}690 hands (10{,}768 frames) from third-person dynamic-gesture videos.
    \item A multidimensional protocol for evaluating complete hand pose estimation pipelines across detection, localization, identification, and temporal stability, complemented by Pose-F1 and stratified analyses across motion-blur classes and hand regions.
    \item A systematic comparison of six off-the-shelf estimators, revealing axis-dependent ranking reversals and estimator-specific failure modes, followed by a targeted post-hoc cleanup case study that demonstrates how the benchmark can validate corrective interventions.
\end{itemize}

\section{Related Work}
\label{sec:related}

\subsection{Hand pose datasets and annotation resources}
Existing hand pose datasets span a broad range of acquisition settings, but their annotations are not uniformly designed for estimator evaluation.
Isolated RGB images are commonly labeled through semi-automatic model fitting \cite{freihand,cocowholebody}; large-scale 3D datasets use multi-view systems or marker-based motion capture in controlled environments \cite{interhand26m,arctic,dexycb,gigahands}; synthetic datasets provide exact annotations at the cost of a domain gap \cite{anyhand,renderih}; and egocentric collections focus primarily on close-range hand--object interaction \cite{assemblyhands,egoexo4d,hot3d}.
As summarized in Table~\ref{tab:datasets}, these resources often rely on model initialization or specialized capture systems, while complete frame-level manual verification remains uncommon.
The per-axis columns make this fragmentation explicit: HInt~\cite{hint}, for example, provides human-corrected 2D hands but targets a designated hand rather than annotating all visible hands exhaustively, and temporal sequences are available only for a subset.

Third-person continuous gestures are instead represented in gesture and sign language corpora \cite{ipnhand,phoenix,isign,how2sign,isharah}.
Although these datasets match our target setting, their released hand landmarks, when available, are typically generated by off-the-shelf estimators without human verification.
For example, How2Sign~\cite{how2sign} uses OpenPose~\cite{openpose}, while iSign~\cite{isign}, Isharah~\cite{isharah}, and the pose extension of IPN Hand~\cite{ipnhandS} rely on MP-holistic~\cite{mp-holistic}.
Their continuous videos support intrinsic temporal analysis and training of recognition models, but the annotations may inherit estimator-specific errors (marked \cmark$^*$ in Table~\ref{tab:datasets}).
Consequently, to the best of our knowledge, no prior third-person resource supports all four evaluation axes under complete human verification, a gap that VERA-Hand fills.

\subsection{Hand pose estimator paradigms}
Current hand pose estimators span several paradigms with distinct representations and system designs.
Hand-specific pipelines predict landmarks from localized hand regions \cite{mp-hands,rtmpose}; whole-body estimators recover the hands jointly with body and facial keypoints \cite{mp-holistic,rtmw}; reconstruction-based methods fit a parametric 3D hand model \cite{mano} and project its joints into the image \cite{wilor}; and human-centric foundation models predict dense whole-body keypoints through large-scale pretraining \cite{sapiens,sapiens2}.

These paradigms also differ in how hands are detected and assigned a hand label.
Although all evaluated pipelines follow a top-down design, the input hand or person region may be obtained via an integrated detector \cite{wilor} or via the default off-the-shelf recommendation provided by the corresponding framework \cite{mmpose}.
Despite these design differences, the resulting systems are commonly used as complete hand pose estimation pipelines in downstream applications, where comparisons often focus solely on keypoint localization.
Their behavior in hand recovery, hand identification, and temporal stability remains less systematically characterized.
VERA-Hand evaluates these properties jointly under a common protocol.

\subsection{Evaluating and benchmarking pose estimators}

A smaller body of work systematically compares pose estimators, following three complementary directions.
First, estimator quality has been evaluated extrinsically via downstream performance in sign language or egocentric action recognition \cite{obrien-slt-eval,assemblyhands}.
Second, metamorphic testing has examined robustness to controlled perturbations such as occlusion, illumination changes, and motion blur, but without human-verified spatial annotations \cite{wang-metamorphic}.
Third, spatial benchmarks report frame-level 2D accuracy using PCK and occlusion labels \cite{hint}, while prior egocentric evaluations separately analyze spatial accuracy and temporal consistency \cite{hands23}.
As summarized in Table~\ref{tab:datasets}, reported evaluations range from task-level accuracy/BLEU to PCK, AP/AR, and MPJPE/PA-MPJPE, and therefore do not necessarily evaluate all dimensions supported by the released annotations.
Ego-Exo4D~\cite{egoexo4d} is the closest resource, providing substantially larger verified multi-view 3D supervision in continuous egocentric video; however, EgoHandPose targets the camera wearer's hands and benchmarks 3D pose using PA-MPJPE and MPJPE.
VERA-Hand complements it with exhaustive frame-level 2D annotations for third-person dynamic gestures and a unified four-axis evaluation protocol covering detection, localization, identification, and temporal stability.

\begin{figure*}[t]
  \centering
  \includegraphics[width=\linewidth]{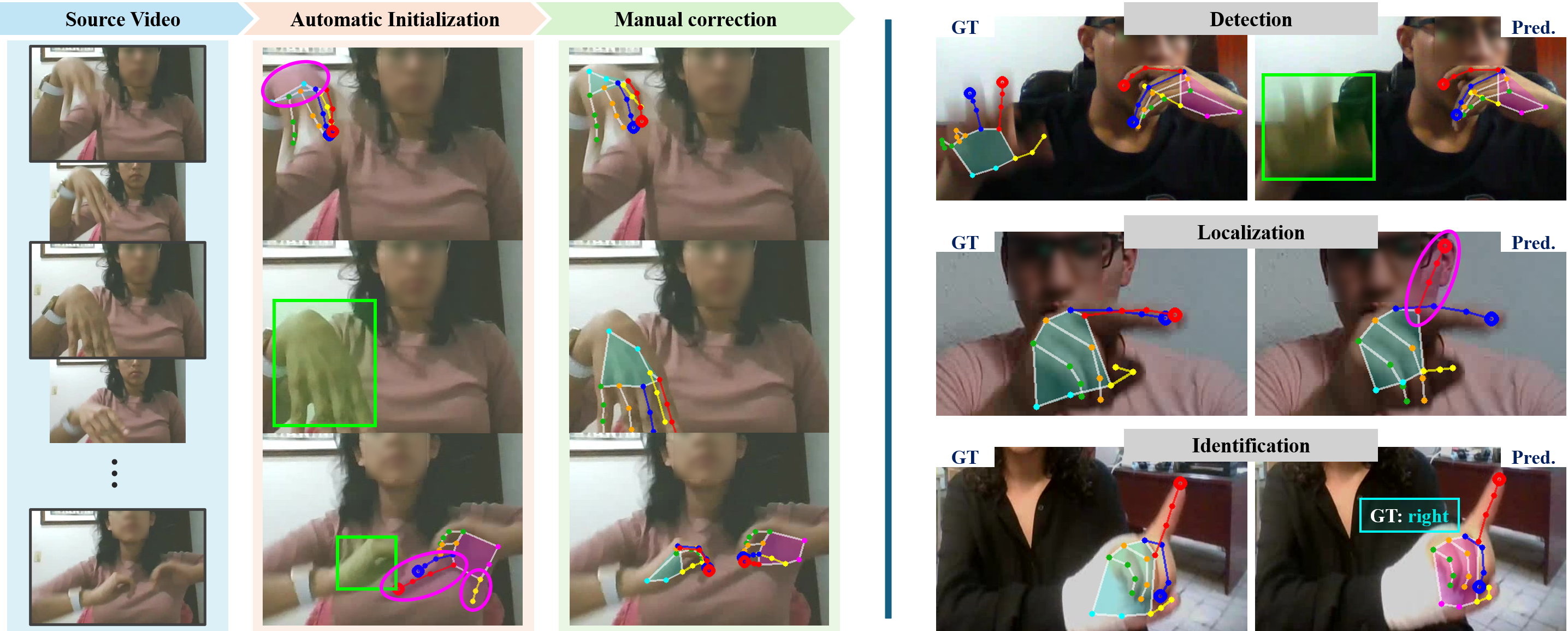}
  \caption{Benchmark construction and evaluation scope. Left: Annotations for sampled frames from gesture videos are automatically initialized and manually corrected. Right: GT and prediction pairs illustrate evaluation of detection, localization, and hand identification.}
  \label{fig:protocol}
\end{figure*}

\section{Benchmark Construction and Protocol}
\label{sec:protocol}

VERA-Hand combines manually verified, frame-level 2D hand pose annotations for third-person continuous dynamic-gesture videos with a multidimensional evaluation protocol (Figure~\ref{fig:protocol}).

\subsection{Source videos and benchmark composition}
\label{subsec:gt}
We build our benchmark on the IPN Hand dataset \cite{ipnhand}, which consists of $640\times480$ RGB videos recorded from a frontal view, with each subject performing continuous gestures to control touchless screens.
Each video comprises roughly twenty gestures, and the hand-to-camera distance varies within a single recording. Fast motion, partial exits from the frame, and hands passing in front of the face occur naturally. 
The benchmark uses the official test split, keeping its annotations disjoint from the dataset's training data, while the source videos remain under its original terms.

Annotation targets approximately 200 keyframes per video, using a category-dependent sampling rate designed to maximize pose variation.
Dynamic gestures are sampled densely (about one frame in two), pointing sparsely (about one frame in seven), and resting at most four frames per instance.
The annotations are therefore not temporally dense, and the protocol is built accordingly: the supervised axes (detection, localization, and identification) are computed on the annotated frames, while the temporal metrics are intrinsic to the predictions and are computed over the full videos without extra annotations.

Each visible hand is annotated independently in every sampled frame with a left/right hand label and 21 two-dimensional keypoints following the MediaPipe topology.
The visibility convention distinguishes two cases: keypoints outside the image are marked as out of frame, whereas occluded keypoints remain annotated, with their locations inferred from hand context.
The annotations were initialized with an off-the-shelf estimator and manually corrected across all sampled frames in two passes conducted several months apart (Figure~\ref{fig:protocol}, left).
A subsequent independent audit was performed to further correct the annotations.

VERA-Hand comprises 10{,}690 annotated hands on 10{,}768 frames from 13 subjects across 52 videos (four per subject).
The data are strongly right-hand-dominant: 10{,}252 frames contain only the right hand, 72 only the left, 183 both, and 261 no hands. The benchmark includes 10{,}435 right-hand and 255 left-hand annotations.
Although smaller than training-oriented corpora, the benchmark is designed for estimator evaluation rather than training.

\subsection{Ground-truth quality and annotation bias}
\label{subsec:quality}
To assess annotation reliability, we independently audited 3{,}500 frames sampled across all 52 videos, focusing on hand presence, handedness, gross pose, and local keypoint placement.
Overall, 268 frames (7.7\%) required corrections; most involved local keypoint adjustments, while five previously missed hands were added. 
All identified issues were corrected in the released annotations, which are the ones used throughout this paper.
A breakdown and representative examples are provided in Appendix~\ref{supp:audit}.

Since the annotations were initialized with MP-holistic \cite{mp-holistic}, its predictions could retain an advantage during evaluation.
We address this potential bias by distinguishing between \emph{biased} and \emph{bias-free} hands.
A hand is classified as biased when the corresponding MP-holistic prediction attains PCK@0.02${=}1$ with respect to the final GT (approximately 3{,}400 hands); all remaining hands form the bias-free subset. Manual correction substantially altered the initialization within this subset: 76\% of the keypoints moved by more than $\alpha{=}0.02$, with a median normalized displacement of 0.06, compared with 0.008 for biased hands.
Moreover, approximately 1{,}381 hands missing from the initialization were added, and 1{,}387 spurious detections were removed from the corresponding frames. Consequently, MP-holistic exhibits a substantial drop in localization performance on the bias-free subset, whereas the other estimators show no comparable agreement with the initialization. 
Unless stated otherwise, all results are reported on the bias-free subset.

\subsection{Evaluation axes and matching}
\label{subsec:axes}
Each axis answers a question the others cannot, and each rests on a precondition the source material must satisfy for that question to be measurable.
All supervised metrics are computed per frame; localization, identification, and Pose-F1 use macro-averaged left/right scores, whereas detection is label-agnostic.

\textbf{Matching.}
Spatial matching uses hand bounding-box IoU between the GT hand and each candidate with a threshold of $0.2$.
Detection uses a global one-to-one, label-agnostic assignment, whereas localization and Pose-F1 are restricted to same-label matching to preserve hand correspondence.

\textbf{Detection} asks whether each GT hand is recovered, independently of its predicted hand label.
Unmatched GT hands are false negatives (Figure~\ref{fig:protocol}, right-top) and unmatched predictions are false positives.
Because the benchmark also includes frames with no hands, false positives are measured in genuinely empty scenes as well as alongside real hands.
We report label-agnostic precision, recall, and F1 (Det-F1), together with two recurrent false-positive patterns:
\emph{duplicate hands} (surplus predictions sharing the same label, either beside a matched hand or without a corresponding GT) and \emph{cross-hand overlap} (opposite-label predictions at nearly the same position, IoU $>0.80$).
Note that detection is evaluated at the system level, using each estimator with the default detector provided by its framework.

\textbf{Localization} asks where the keypoints fall once the corresponding hand is found and is computed over matched true positives only.
We report PCK at $\alpha{=}0.10$ and MPJPE, normalized by hand size (wrist-to-middle-MCP distance).
Out-of-frame GT keypoints are excluded from the metric, while occluded keypoints are scored normally.

\textbf{Identification} asks whether each detected hand is assigned the correct hand label.
We measure it through the label-swap rate using a label-agnostic spatial match: the highest-IoU candidate above $0.2$ is selected regardless of label, and a swap is counted when the labels disagree (Figure~\ref{fig:protocol}, right-bottom).
Cross-hand overlap cases in which predictions of both hand labels overlap the same GT hand are excluded from this metric.
The resulting swap rate, therefore, captures unambiguous hand label flips without being inflated by overlapping duplicate predictions.

\textbf{Temporal stability} asks how steady the predictions remain over time.
This axis is intrinsic, since it uses no temporal GT and characterizes prediction dynamics over the full videos rather than temporal accuracy.
Following \cite{obrien-slt-eval}, we compute acceleration- and jerk-based jitter, $J_{\text{acc}}$ and $J_{\text{jerk}}$, as the mean magnitudes of the second- and third-order temporal differences, respectively. We additionally report motion energy, $E_v$ (mean joint velocity), to contextualize smoothness relative to the predicted motion.
All metrics are scaled by 100 and computed from image-normalized coordinates for three keypoint subsets: all 21 keypoints, the base joints (wrist and MCPs), and the fingertips. We also count tracking gaps, defined as runs of consecutive missed detections bounded by detections of the same hand.
Detailed keypoint-subset results and tracking-gap analysis are provided in Appendix~\ref{supp:temporal}.

\subsection{Pose-F1: Joint Detection and Localization}
\label{subsec:posef1}
Detection and localization can be misleading when considered independently.
A pipeline may achieve high detection recall by producing excessive predictions, while localization, computed only over matched hands, conditions evaluation on successful detection.
Pose-F1 evaluates both requirements jointly.
A prediction is counted as a true positive only if it is matched to a GT hand under same-label matching and satisfies $\mathrm{PCK}@\alpha \geq \beta$.
A matched prediction that fails this criterion is counted as a false positive, with its corresponding GT hand as a false negative;
unmatched predictions are counted as false positives and unmatched GT hands as false negatives, following the same accounting as detection.
Precision, recall, and F1 (Pose-F1) are then computed from these counts.
We report $\alpha{=}0.10$ and $\beta{=}0.5$ in the main paper.
Since neither permissive detection nor accurate localization alone can raise Pose-F1, we use it to assess whether the case study in Section~\ref{subsec:cleanup} produces joint improvements.

\subsection{Stratified reporting}
\label{subsec:strat}
Aggregate scores indicate whether an estimator fails, whereas stratified results reveal under which conditions and in which hand regions those failures occur. We therefore report performance under two complementary stratifications.
For \textbf{motion blur}, each hand instance is assigned to one of three classes using two signals obtained within the hand region: the 90th-percentile magnitude of Farneback optical flow~\cite{farneback} ($\mathrm{F}_{p90}$), which captures hand motion, and the variance of the Laplacian ($\mathrm{L}_{\mathrm{var}}$), which measures local sharpness. An instance is classified as follows:
\begin{equation}
\text{blur class} =
\begin{cases}
\text{blurry} &
\begin{aligned}
&\text{if } \mathrm{F}_{p90} > 13.0 \text{ or} \\
&(\mathrm{F}_{p90} > 6.5
  \text{ and } \mathrm{L}_{\mathrm{var}} < 31.0),
\end{aligned}
\\[3pt]
\text{sharp} & \text{if } \mathrm{F}_{p90} \leq 7.5 \text{ and } \mathrm{L}_{\mathrm{var}} > 55.0,
\\[3pt]
\text{medium} & \text{otherwise}.
\end{cases}
\label{eq:blur_classification}
\end{equation}
Thresholds were calibrated by visual inspection, and examples from each class are provided in Appendix~\ref{supp:blur}.
For \textbf{hand region}, localization is reported separately for the wrist, palm joints, and fingertips, as well as over all 21 keypoints. This stratification reveals which anatomical regions most strongly differentiate the estimators.

\begin{table}[t]
  \centering
  \footnotesize
  \setlength{\tabcolsep}{3.5pt}
  \renewcommand{\arraystretch}{1.15}
  \begin{tabular}{l l l c}
    \toprule
    Estimator & Paradigm & Detector & Native kpts \\
    \midrule
    MP-hands \cite{mp-hands}       & Hand-specific  & BlazePalm \cite{mp-hands}               & 21       \\
    RTMP-hand \cite{rtmpose}       & Hand-specific  & Cascade R-CNN \cite{cascadercnn} & 21     \\
    MP-holistic \cite{mp-holistic} & Whole-body     & BlazePose \cite{blazepose}     & 543 (wb) \\
    RTMW-wb \cite{rtmw}            & Whole-body     & RTMDet \cite{rtmdet}           & 133 (wb) \\
    WiLoR \cite{wilor}             & Reconstruction & Custom YOLO \cite{wilor}            & 21 (3D)  \\
    Sapiens2-1B \cite{sapiens2}    & Foundation     & DETR \cite{detr}               & 308 (wb) \\
    \bottomrule
  \end{tabular}
  \caption{Evaluated estimators, paradigms, and default detectors. Native kpts denote each model's output before mapping.}
  \label{tab:estimators}
\end{table}

\subsection{Estimators and evaluation setup}
\label{sec:setup}
The six evaluated estimators summarized in Table~\ref{tab:estimators} cover the four paradigms discussed in Section~\ref{sec:related}.
To preserve off-the-shelf comparability, each pipeline uses its official default configuration, confidence thresholds, and detector, without dataset-specific tuning. 
Detection is therefore assessed as part of the complete deployed pipeline rather than independently of the pose estimator.
The main comparison is based on the raw predictions, while any post-processing intervention, including the cleanup examined in Section~\ref{subsec:cleanup}, is evaluated separately.
Native outputs are mapped to a common topology of 21 two-dimensional keypoints.
This includes projecting WiLoR's 3D MANO \cite{mano} joints onto the image plane and applying the Goliath-to-21 mapping for the Sapiens2 layout.
Additional conversion details are provided in Appendix~\ref{supp:setup}.

\section{Results and Analysis}
\label{sec:results}

\subsection{No single estimator dominates}
\label{subsec:flip}
The estimator ranking changes substantially across evaluation axes (Table~\ref{tab:main}, Figure~\ref{fig:teaser}).
WiLoR achieves the highest detection (Det-F1: 0.978) and the lowest temporal jitter ($J_{\text{jerk}}$: 1.181), yet ranks second-last in localization (PCK: 0.47). MP-hands exhibits the opposite pattern: it provides the highest localization (0.573), while attaining the largest label-swap rate (4.41\%). Sapiens2 performs worst in detection (0.666) but best on identification (0.03\%, the lowest rate). 
Simultaneously, Pose-F1 yields yet another ranking led by RTMW-wb (0.477), despite not leading in either detection or localization individually. 
Overall, four estimators account for the five metric-wise best results, demonstrating that no single score captures the different dimensions of estimator reliability.
These rankings are not an artifact of benchmark size: subsampling to 75\%, 50\%, 25\%, and 10\% of the frames preserves the relative ordering on all four metrics, with detection varying by less than 0.01 and the Sapiens2 cross-hand overlap remaining above 82\% (Appendix~\ref{supp:subsampling}).

To quantify the effect of the MP-holistic annotation initialization, we additionally evaluate it on the full annotated set. Its PCK increases from 0.439 on the bias-free subset to 0.579 on the full set, confirming the localization advantage introduced by the initialization and supporting the bias-controlled protocol used throughout the main comparison. Further bias results are provided in Appendix~\ref{supp:bias}.

\begin{table}[t] 
  \centering 
  \footnotesize 
  \setlength{\tabcolsep}{4.5pt} 
  \renewcommand{\arraystretch}{1.15} 
  \begin{tabular}{l ccccc} 
    \toprule 
    Estimator & Det-F1 $\uparrow$ & PCK $\uparrow$ & Pose-F1 $\uparrow$ & Swap $\downarrow$ & $J_{\text{jerk}}$ $\downarrow$ ($E_v$) \\ 
    \midrule 
    MP-holistic & 0.840 & 0.439 & 0.309 & 0.51\% & 1.59 (1.52) \\ 
    MP-hands    & 0.898 & \textbf{0.573} & 0.389 & 4.41\% & 1.69 (1.51) \\ 
    WiLoR       & \textbf{0.978} & 0.470 & 0.407 & 0.11\% & \textbf{1.18} (1.38) \\ 
    RTMW-wb     & 0.932 & 0.560 & \textbf{0.477} & 0.31\% & 1.47 (1.38) \\ 
    RTMPhand   & 0.872 & 0.489 & 0.340 & 3.51\% & 1.76 (1.43) \\ 
    Sapiens21B & 0.666 & 0.491 & 0.361 & \textbf{0.03\%} & 1.30 (1.39) \\ 
    \bottomrule 
  \end{tabular} 
  \caption{Comparison across detection (Det-F1), localization (PCK@0.10), the joint Pose-F1@0.10$\geq$0.5, identification (label-swap rate), and temporal stability ($J_{\text{jerk}}$, with motion energy $E_v$ in parentheses).
  } 
  \label{tab:main} 
\end{table}

\subsection{Per-dimension findings}
\label{subsec:perdim}
We next examine each evaluation axis individually. \textbf{Detection} performance varies substantially, with Det-F1 ranging from 0.666 to 0.978, not following a clear hand-specific versus whole-body distinction.
WiLoR, which uses a dedicated hand detector, and RTMW-wb, which relies on person detection, achieve the two highest scores despite their different detection strategies. Thus, the gap one might expect between hand-level and person-level detection does not clearly materialize in this setting. WiLoR leads by a considerable margin, whereas Sapiens2 obtains substantially lower Det-F1. Detection errors also include spurious predictions, as analyzed in Section~\ref{subsec:cleanup}.
\textbf{Localization} reverses this ranking. Among correctly matched hands, MP-hands achieves the highest PCK (0.573), whereas WiLoR, the detection leader, obtains the second-lowest (0.47). Because WiLoR predictions are mapped to 2D keypoints by projecting a reconstructed 3D representation \cite{mano} rather than directly regressing image-space landmarks, this result may reflect a trade-off between robust hand recovery and precise 2D localization.
More importantly, this ranking inversion demonstrates the need for VERA-Hand's multidimensional evaluation: a localization-only comparison would poorly rank WiLoR while overlooking its leading performance in detection and temporal stability.

\textbf{Pose-F1} makes the joint reading explicit, crediting a hand only when it is both detected and accurately localized.
Here RTMW-wb achieves the highest performance (0.477) by combining strong detection (0.932) with competitive localization (0.56), despite leading neither axis individually.
In contrast, MP-hands' localization advantage is limited by its detection performance, yielding a Pose-F1 of 0.389, while WiLoR drops from first in detection to second in Pose-F1. Thus, the joint metric favors balanced pipelines over estimators that excel along only one dimension.

\textbf{Identification} reveals a marked difference between hand-specific models and the rest.
MP-hands and RTMP-hand exhibit label-swap rates of 4.41\% and 3.51\%, respectively, whereas all other estimators remain at or below 0.51\%.
For the whole-body and foundation models, explicit body context may help disambiguate left from right.
WiLoR, despite operating on individual hands, achieves a comparably low rate (0.11\%), possibly aided by its custom YOLO detector and handedness-assignment mechanism.
On the other hand, \textbf{temporal stability} does not track spatial quality and can even oppose it.
WiLoR has the lowest jitter ($J_{\text{jerk}}$: 1.18) despite relatively weak localization, while MP-hands and RTMP-hand, which perform comparatively well in localization, exhibit the highest jitter values (1.69 and 1.76).
Because low jitter can also reflect limited predicted motion, Table~\ref{tab:main} also reports $E_v$ alongside $J_{\text{jerk}}$; the detailed analysis is provided in Appendix~\ref{supp:temporal}.
Thus, these metrics characterize prediction smoothness rather than correctness and should not be interpreted in isolation.
Qualitative comparisons between the GT and the estimator's predictions are provided in Appendix~\ref{supp:qualitative}.

\begin{figure}[t]
  \centering
  \includegraphics[width=\linewidth]{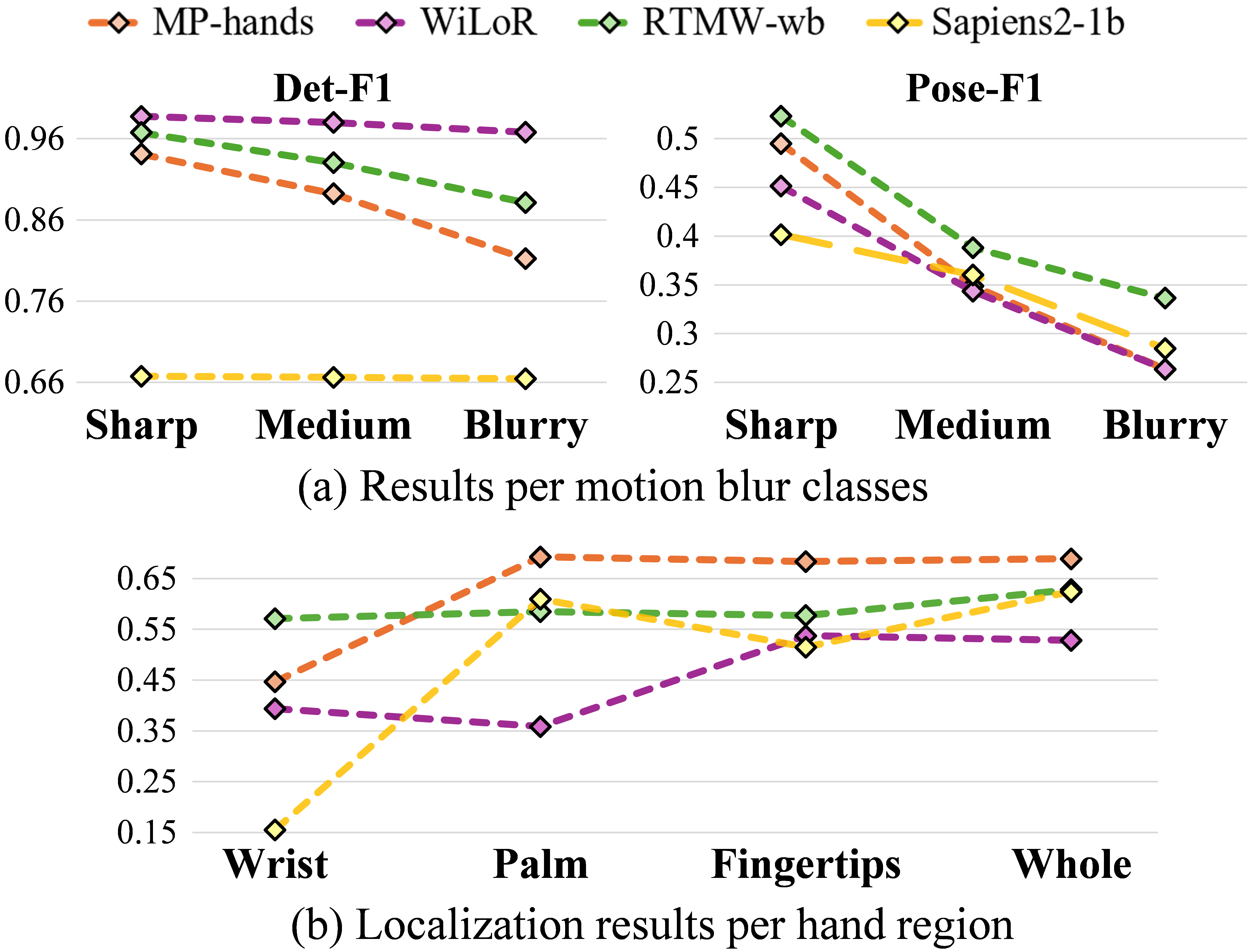}
  \caption{Stratified analysis for four representative estimators (full set in Appendix~\ref{supp:results}). (a) Det-F1 and Pose-F1 across motion-blur classes. (b) Right-hand PCK across hand regions.}
  \label{fig:blur}
\end{figure}

\subsection{Stratified analysis: blur and region}
\label{subsec:strat-results}
Stratification by motion blur shows that robustness depends on the evaluation dimension (Figure~\ref{fig:blur}a).
Detection remains highly stable for WiLoR and Sapiens2, while the hand-specific estimators are more sensitive to blur: MP-hands decreases from 0.94 to 0.81, with RTMP-hand showing a similar trend in Appendix~\ref{supp:blur}.
Pose-F1 decreases from sharp to blurry for all estimators.
RTMW-wb leads across all three classes, although its score drops from 0.57 to 0.39.
MP-hands is competitive on sharp instances (0.49) but falls to 0.27 under blur, similar to RTMP-hand and WiLoR. 
This reveals that no single estimator is uniformly robust to blur across evaluation axes, just as none dominates the aggregate ranking.

Localization errors also vary significantly across hand regions (Figure~\ref{fig:blur}b).
We focus on the right hand, which is substantially better represented in VERA-Hand, and compute PCK only over correctly matched predictions to isolate localization from detection.
The wrist shows the largest separation among estimators, ranging from 0.16 to 0.57, while fingertips range from 0.52 to 0.68. The pattern is strongly estimator-dependent: Sapiens2 is the worst at the wrist (0.16) while remaining competitive over the palm and whole hand, whereas WiLoR is relatively weak at the wrist (0.39) and weakest at the palm (0.36). In this setting, the wrist, rather than the fingertips, is the region that most clearly distinguishes estimator behavior.

\begin{table}[t]
  \centering
  \footnotesize
  \setlength{\tabcolsep}{4pt}
  \renewcommand{\arraystretch}{1.15}
  \begin{tabular}{l cc cc cc}
    \toprule
    & \multicolumn{2}{c}{Cross-hand overlap} & \multicolumn{2}{c}{Det-F1} & \multicolumn{2}{c}{Pose-F1} \\
    \cmidrule(lr){2-3}\cmidrule(lr){4-5}\cmidrule(lr){6-7}
    Estimator & raw & clean & raw & clean & raw & clean \\
    \midrule
    MP-holistic & 4.82\%  & 3.32\% & 0.840 & \textbf{0.888} & 0.309 & 0.325 \\
    MP-hands    & 0\%  & 0\% & 0.898 & 0.926          & 0.389 & \textbf{0.442} \\
    WiLoR       & 0.07\%  & 0.03\% & 0.978 & 0.987          & 0.407 & 0.419 \\
    RTMW-wb     & 0.91\%  & 0.03\% & 0.932 & 0.951          & 0.477 & 0.477 \\
    RTMP-hand   & 0.02\%  & 0.09\% & 0.872 & 0.883          & 0.340 & 0.342 \\
    Sapiens2-1B & 85.44\% & 0.05\% & 0.666 & \textbf{0.940} & 0.361 & \textbf{0.406} \\
    \bottomrule
  \end{tabular}
  \caption{Effect of post-hoc cleanup on all six estimators. Results before (\emph{raw}) and after (\emph{clean}) cleanup. Bold values indicate the two largest improvements within Det-F1 and Pose-F1.}
  \label{tab:cleanup}
\end{table}

\subsection{Cross-hand overlap and post-hoc cleanup}
\label{subsec:cleanup}
Sapiens2 predicts a second hand with the opposite label over the real one in 85.44\% of all benchmark frames (Table~\ref{tab:cleanup}).
This cross-hand overlap rate is more than an order of magnitude higher than that of any other estimator (all below 5\%). The failure also becomes more frequent across Sapiens generations: Sapiens1 \cite{sapiens} exhibits cross-hand overlap in 14.1\% of frames, despite Sapiens2 providing better PCK and label-swap performance.
The overlap occurs mostly when only one hand is present and nearly disappears when both hands are visible.
This suggests that Sapiens hallucinates the absent hand rather than missing the visible one, possibly due to its full-body training setting.
VERA-Hand exposes this failure through its predominantly single-hand setup, in which opposite-handed phantom predictions are directly observable.
By contrast, the Sapiens family may appear robust in downstream evaluations on videos where both hands are consistently present, because it rarely drops either hand \cite{obrien-slt-eval};
here, the same persistence becomes a severe source of false positives and opposite-handed phantom predictions when one hand is genuinely absent.
Additional comparisons with Sapiens1 and breakdowns are provided in Appendix~\ref{supp:sapiens}.

As a case study, we use the benchmark to evaluate a post-hoc, training-free cleanup applied to the predictions.
It comprises two stages that address distinct failure modes: a spatial stage resolves cross-hand overlap by retaining the higher-confidence hand when opposite-handed predictions coincide, while a temporal stage uses a short frame window to remove duplicates and fill brief detection gaps via interpolation.
The cleanup benefits different estimators (Table~\ref{tab:cleanup}).
For Sapiens2, cleanup nearly eliminates cross-hand overlap (85.44\% $\rightarrow$ 0.05\%) and produces by far the largest Det-F1 gain from 0.666 to 0.94, pushing the rank from last to third.
MP-holistic shows the second-largest detection gain, while the remaining estimators change only modestly.
Pose-F1 gains are smaller: MP-hands improves the most (0.389 $\rightarrow$ 0.442), overtaking WiLoR (0.419), whereas RTMW-wb remains highest (0.477).
Taken together, these results show how the protocol links distinct failure patterns to targeted improvements across evaluation dimensions.
VERA-Hand therefore provides more than an estimator ranking: it helps identify the failure mechanisms underlying the scores and evaluate targeted corrective interventions.

\section{Discussion and Limitations}
\label{sec:discussion}

The ranking reversals across evaluation axes have a direct practical implication: asking which estimator is best is incomplete unless the reliability dimension relevant to the application is specified.
An application that relies on consistent hand identity over time may favor a different pipeline from one that depends on precise keypoint localization, and an estimator that is excellent for one can be a poor choice for the other.
VERA-Hand makes these trade-offs actionable by reporting reliability along complementary axes rather than collapsing it into a single score.

This application-dependent view also clarifies the relationship between component-level and downstream evaluation. 
Downstream evaluation remains essential, but it measures end-task performance and may overlook failures of the pose estimator itself that become apparent when evaluated against a verified 2D reference \cite{obrien-slt-eval}.
VERA-Hand is therefore complementary to task-level evaluation, not a replacement for it.
The post-hoc cleanup case study illustrates this role: its contribution lies not in the cleanup procedure itself, but in demonstrating that a multidimensional protocol can identify a specific failure mode, determine which reliability axes it affects, and assess whether a corrective intervention improves the targeted behavior while monitoring potential regressions elsewhere.

VERA-Hand has some limitations that should be considered when interpreting its findings.
First, it covers a single regime (the third-person dynamic gestures of IPN Hand), so the observed rankings should not be assumed to transfer to egocentric, close-range, or other domains.
Extending the protocol to additional corpora remains important. At the same time, subsampling shows that the main comparative conclusions remain stable even with substantially fewer annotated frames (Appendix~\ref{supp:subsampling}).
Second, the dataset is strongly right-hand-dominant, containing only 255 left hands.
For the label-aware axes, macro averaging prevents right-hand performance from dominating the aggregate scores, but the limited left-hand annotations make left-hand estimates less conclusive.
Finally, VERA-Hand evaluates 2D rather than 3D pose accuracy, and the annotations were initialized from an existing estimator and then manually corrected. 
The independent audit identified issues in 7.7\% of the reviewed frames, all of which were corrected; nevertheless, some residual annotation errors and initialization biases may persist beyond the audited sample.

Two further limitations concern the scope of the off-the-shelf evaluation. 
The main benchmark evaluates only the Sapiens2-1B checkpoint with its official weights.
Although the results for the preceding Sapiens generation in Appendix~\ref{supp:sapiens} provide an initial point of comparison, broader testing across model sizes and checkpoints is needed to determine how consistently the observed cross-hand overlap occurs across the family.
More generally, each estimator was evaluated as a complete released pipeline using its default detector and pose configuration.
This reflects how practitioners commonly deploy these systems, but it does not disentangle detector errors from pose-estimation behavior or identify the best detector--pose combinations, which would require a dedicated factorial study.
Accordingly, the reported rankings should be interpreted as comparisons of complete released pipelines rather than isolated pose heads.

\section{Conclusion}
\label{sec:conclusion}

We introduced VERA-Hand, a manually verified, frame-level 2D hand pose benchmark for third-person continuous dynamic gestures, along with a protocol that evaluates detection, localization, identification, and temporal stability, including Pose-F1 as a joint metric of detection and localization.
Our comparison of six off-the-shelf estimators shows that no single pipeline dominates across evaluation axes: rankings change substantially from one dimension to another, demonstrating that localization alone provides an incomplete account of reliability.
VERA-Hand also reveals estimator-specific failure modes that aggregate or localization-only evaluations may overlook, including cross-hand overlap.
At the same time, the post-hoc cleanup case study demonstrates how the protocol can assess targeted corrective interventions and monitor their effects across reliability axes. 
We release the annotations, protocol, and code to support reproducible multidimensional evaluation of hand pose estimators.

{
    \small
    \bibliographystyle{ieeenat_fullname}
    \bibliography{main}
}

\clearpage
\appendix

\section*{Supplementary Material}

This supplementary material provides methodological details and extended results that support the main paper.
Section~\ref{supp:audit} reports the annotation audit and characterizes the audited sample.
Section~\ref{supp:setup} documents the estimator configurations and keypoint conversions required for reproducibility.
Section~\ref{supp:results} provides the complete motion-blur, hand-region, temporal-stability, annotation-bias, and benchmark size sensitivity analysis.
Section~\ref{supp:sapiens} further characterizes the Sapiens family and the conditions under which cross-hand overlap occurs.
Section~\ref{supp:cleanup} specifies the post-hoc cleanup procedure and presents its ablation.
Section~\ref{supp:qualitative} presents qualitative comparisons between GT and the estimator predictions.
Unless otherwise stated, supervised results use the bias-free subset; localization, identification, and Pose-F1 are macro-averaged over the left and right hands, whereas detection is label-agnostic. Temporal analyses are computed over the complete videos.

\section{Annotation Audit}
\label{supp:audit}

To assess the reliability of the manually verified annotations, we conducted two independent audits, each carried out by a reviewer who did not participate in the keypoint annotation. The first inspected 500 frames and the second a further 3{,}000, for a total of 3{,}500 audited frames spanning all 13 subjects and all 52 videos. Each frame was examined for four categories of error: hand presence, handedness label, gross pose configuration, and local keypoint placement. The audits were corrective rather than diagnostic: every issue identified in either pass was fixed, and the corrected annotations are the ones released and used throughout the paper.

The audited set is deliberately not a proportional sample (Table~\ref{tab:audit}). Because left and bimanual hands are rare in VERA-Hand, the audit covers all 72 left-only frames and all 183 bimanual frames, together with roughly one third of the right-only frames. Left hands therefore account for 6.9\% of the audited hands against 2.4\% of the benchmark, so the rare and most error-prone configurations are inspected exhaustively rather than in proportion. Coverage of out-of-frame keypoints (12.5\% of audited frames against 14.1\% of the benchmark) and of the motion-blur classes remains close to the benchmark composition.
The 261 hand-free frames fall outside this audit, since they contain no keypoint annotations for inspection; each was confirmed to contain no visible hand at selection time.

\begin{figure}[t]
  \centering
  \includegraphics[width=0.8\linewidth]{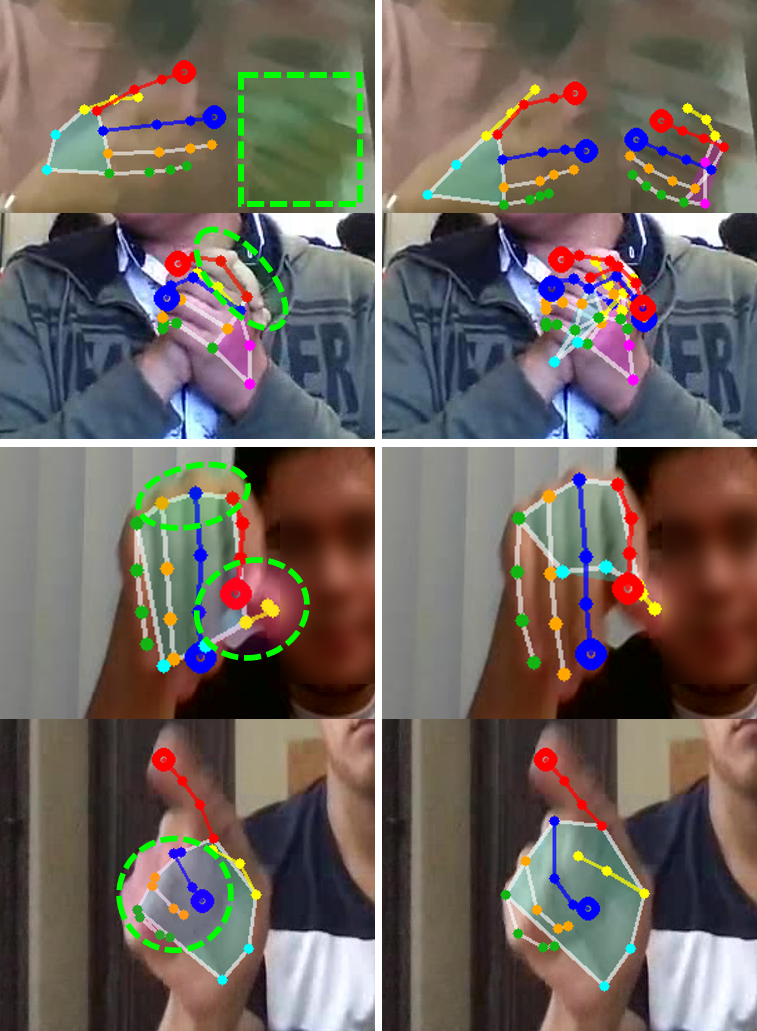}
  \caption{Representative corrections from the audit. Each row shows one frame before (left) and after (right) correction, with green markers indicating the affected region. The top two rows show previously missed hands that were added; the remaining rows show local keypoint displacements.}
  \label{fig:audit_examples}
\end{figure}

Across the 3{,}500 audited frames, 268 (7.7\%) required a correction. Of these, 263 were small displacements of individual keypoints and five were previously missed hands that were added; no errors were found in handedness labels or gross pose configuration. Residual uncertainty is therefore concentrated in fine-grained localization rather than in hand-level validity, and the two passes agree on its scale (7.2\% and 7.7\% of their respective frames). Figure~\ref{fig:audit_examples} shows representative corrections.

\begin{table}[t]
  \centering
  \footnotesize
  \setlength{\tabcolsep}{3.7pt}
  \renewcommand{\arraystretch}{1.2}
  \begin{tabular}{l cc}
    \toprule
    Property & Audited (3{,}500) & Full benchmark \\
    \midrule
    Subjects / Videos            & 13 / 52                 & 13 / 52 \\
    Frames (right / left / both) & 3{,}245 / 72 / 183      & 10{,}252 / 72 / 183 \\
    Frames without hands         & 0                       & 261 \\
    Hands (right / left)         & 3{,}428 / 255           & 10{,}435 / 255 \\
    Frames w/out-of-frame kpts  & 437                     & 1{,}477 \\
    Blur (S / M / B)             & 1{,}447 / 1{,}105 / 948 & 4{,}338 / 3{,}379 / 2{,}790 \\
    \bottomrule
  \end{tabular}
  \caption{Composition of the audited set relative to the full benchmark, where S, M, and B denote the sharp, medium, and blurry classes. All left-only and bimanual frames are audited, so the rare configurations are covered exhaustively rather than proportionally. Blur is counted per frame over the frames containing hands, using the right-hand class for bimanual frames; this convention is used only for this descriptive comparison, since all results in the paper are computed per hand.}
  \label{tab:audit}
\end{table}

\section{Estimator Setup and Keypoint Conversion}
\label{supp:setup}

Each estimator is evaluated using its official off-the-shelf configuration, including the default detector, confidence thresholds, preprocessing, and input resolution, without dataset-specific tuning.
The MMPose pipelines~\cite{mmpose} use the RTMPose-m Hand5 checkpoint for RTMP-hand~\cite{rtmpose} at $256{\times}256$ and the RTMW-x Cocktail14 checkpoint for RTMW-wb~\cite{rtmw} at $384{\times}288$.
Sapiens2~\cite{sapiens2} uses the 1B Goliath pose checkpoint at $1024{\times}768$, while MP-hands~\cite{mp-hands} and MP-holistic~\cite{mp-holistic} use MediaPipe release 0.10.32.
WiLoR~\cite{wilor} is evaluated using its publicly released checkpoint.
The detectors and per-model details are summarized in Table~\ref{tab:estimators}.

Handedness is obtained differently across pipelines, which is relevant to the identification results.
MP-holistic, RTMW-wb, and Sapiens2 encode the left and right hands using fixed output keypoint indices.
MP-hands provides a handedness label through its native classifier, whereas WiLoR uses the side class predicted by its YOLO-based hand detector.
RTMP-hand is the only evaluated pipeline whose detector returns hand bounding boxes without a side label, requiring handedness to be assigned post hoc.
For this purpose, we use RTMPose-L-WholeBody from the same MMPose framework, with a $384{\times}288$ input and the same RTMDet~\cite{rtmdet} person detector used for RTMW-wb.
This auxiliary pipeline is used exclusively to obtain left- and right-wrist anchors and does not alter the hand keypoints predicted by RTMP-hand.
Each RTMP-hand detection is assigned the label of the nearest compatible wrist anchor.
Anchors located inside the hand box are preferred, followed by anchors lying within one box diagonal of its center.
When no suitable anchor is available, handedness is inferred from the predicted hand geometry using a palm cross-product.
Because RTMP-hand relies on an inferred rather than native handedness label, this additional assignment step may contribute to its comparatively high label-swap rate.

All native outputs are mapped to the common 21-keypoint MediaPipe topology.
Most mappings require only index selection, but Sapiens2 and WiLoR require additional transformations.
In the Sapiens2 Goliath layout, each finger is ordered from tip to base, and the wrist follows the finger joints; we reverse each finger block to the MediaPipe base-to-tip order and move the wrist to the first index.
WiLoR predicts 21 MANO~\cite{mano} joints in 3D, which are projected onto the image plane using the model-predicted camera.
For left hands, the projected coordinates are mirrored horizontally to match the annotation convention.

\section{Extended Quantitative Results}
\label{supp:results}

This section extends the analyses summarized in the main paper by reporting complete results for all six estimators.
We first examine performance across motion-blur classes and hand regions, followed by detailed temporal-stability results and an analysis of initialization bias.

\begin{figure}[t]
  \centering
  \includegraphics[width=0.9\linewidth]{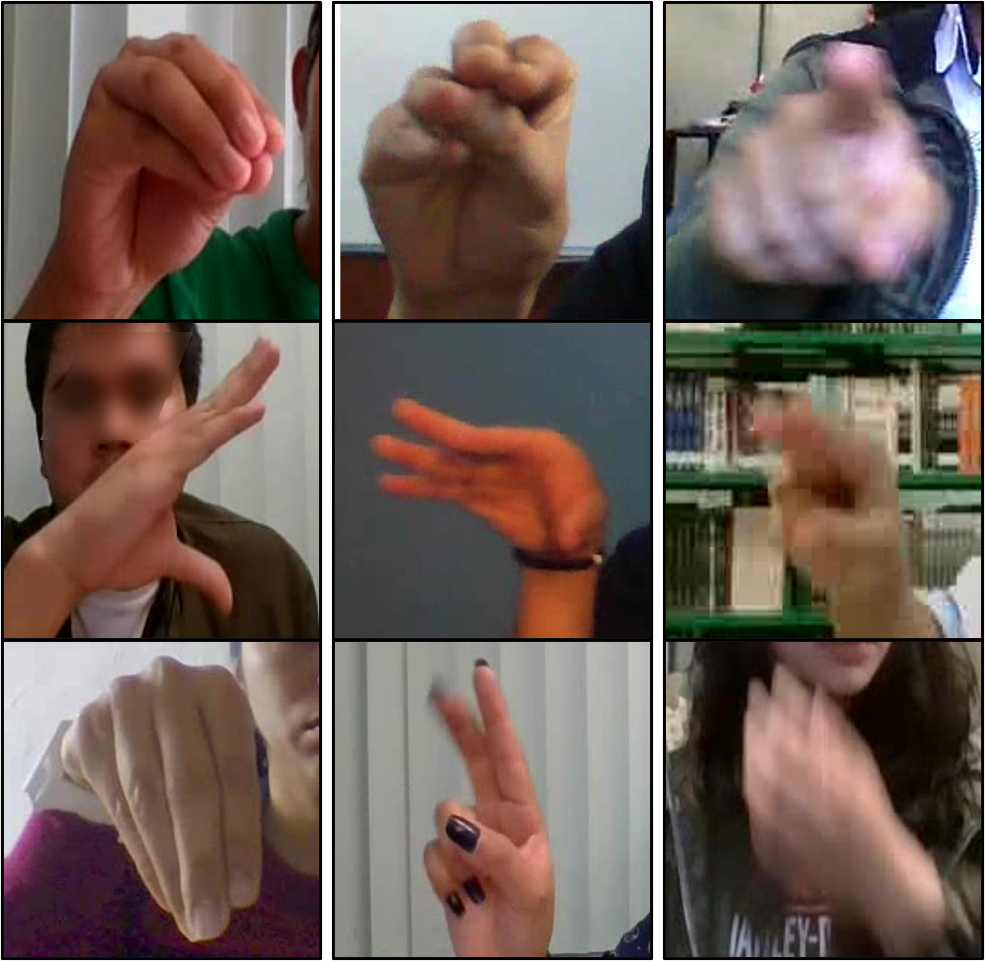}
  \caption{Representative hand crops from the three motion-blur classes, one class per column (left: sharp; middle: medium; right: blurry). Blur is assigned per hand from optical-flow magnitude and hand-region sharpness.}
  \label{fig:blur_examples}
\end{figure}

\subsection{Motion blur}
\label{supp:blur}

Table~\ref{tab:blur} reports the motion-blur stratification for all six estimators, extending the four representative pipelines shown in Figure~\ref{fig:blur}.
Representative examples of the sharp, medium, and blurry classes are shown in Figure~\ref{fig:blur_examples}.
Blur is assigned independently to each annotated hand, whereas the frame-level assignment in Table~\ref{tab:audit} is used only to describe the audited sample.
All supervised results in this analysis are therefore computed at the hand level.

Detection is nearly invariant to blur for MP-holistic, Sapiens2-1B, and WiLoR, whereas the hand-specific pipelines degrade the most (RTMP-hand drops by 0.19 and MP-hands by 0.13) and RTMW-wb shows an intermediate decline. PCK and Pose-F1 degrade for all estimators and the rankings cross.
Identification is largely unaffected by blur, and label-swap rates show no common blur-dependent trend.
These results show that robustness to motion blur depends on both the estimator and the evaluation axis, with no pipeline remaining uniformly robust across all metrics.

\begin{table*}[t]
  \centering
  \footnotesize
  \setlength{\tabcolsep}{8pt}
  \renewcommand{\arraystretch}{1.2}
  \begin{tabular}{l ccc ccc ccc ccc}
    \toprule
    & \multicolumn{3}{c}{Det-F1 $\uparrow$} & \multicolumn{3}{c}{PCK $\uparrow$} & \multicolumn{3}{c}{Pose-F1 $\uparrow$} & \multicolumn{3}{c}{Swap $\downarrow$} \\
    \cmidrule(lr){2-4}\cmidrule(lr){5-7}\cmidrule(lr){8-10}\cmidrule(lr){11-13}
    Estimator & S & M & B & S & M & B & S & M & B & S & M & B \\
    \midrule
    MP-holistic & 0.841 & 0.835 & 0.845 & 0.485 & 0.424 & 0.397 & 0.356 & 0.303 & 0.276 & 0.57\% & 0.53\% & 0.42\% \\
    MP-hands    & 0.941 & 0.892 & 0.812 & 0.654 & 0.523 & 0.452 & 0.494 & 0.350 & 0.267 & 3.28\% & 4.98\% & 5.49\% \\
    WiLoR       & 0.987 & 0.980 & 0.968 & 0.520 & 0.469 & 0.402 & 0.506 & 0.394 & 0.307 & 0.11\% & 0.09\% & 0.14\% \\
    RTMW-wb     & 0.968 & 0.931 & 0.881 & 0.634 & 0.542 & 0.463 & 0.568 & 0.431 & 0.390 & 0.25\% & 0.41\% & 0.28\% \\
    RTMP-hand   & 0.938 & 0.873 & 0.751 & 0.532 & 0.482 & 0.409 & 0.418 & 0.328 & 0.218 & 3.99\% & 3.66\% & 2.58\% \\
    Sapiens2-1B & 0.668 & 0.666 & 0.664 & 0.554 & 0.477 & 0.413 & 0.407 & 0.363 & 0.290 & 0.05\% & 0.00\% & 0.03\% \\
    \bottomrule
  \end{tabular}
  \caption{Performance across motion-blur classes for all six estimators. S, M, and B denote sharp, medium, and blurry instances, respectively. PCK and Pose-F1 decrease consistently with blur, whereas detection and label-swap rates exhibit estimator-dependent patterns. Results use per-hand blur assignments and the bias-free subset; localization, Pose-F1, and identification are macro-averaged, whereas detection is label-agnostic. PCK is evaluated at $\alpha{=}0.10$, and Pose-F1 uses $\alpha{=}0.10$ and $\beta{=}0.5$.}
  \label{tab:blur}
\end{table*}

\begin{table}[t]
  \centering
  \footnotesize
  \setlength{\tabcolsep}{6pt}
  \renewcommand{\arraystretch}{1.15}
  \begin{tabular}{l cccc}
    \toprule
    Estimator & Wrist & Palm & Fingertips & Whole \\
    \midrule
    MP-holistic & 0.552 & 0.622 & 0.607 & 0.612 \\
    MP-hands    & 0.446 & 0.691 & 0.683 & 0.688 \\
    WiLoR       & 0.394 & 0.359 & 0.538 & 0.529 \\
    RTMW-wb     & 0.571 & 0.585 & 0.577 & 0.628 \\
    RTMP-hand   & 0.346 & 0.546 & 0.552 & 0.611 \\
    Sapiens2-1B & 0.155 & 0.609 & 0.515 & 0.624 \\
    \bottomrule
  \end{tabular}
  \caption{PCK@0.10 by hand region for the right hand. Localization is computed over correctly matched predictions. The wrist exhibits the largest cross-estimator spread.}
  \label{tab:region}
\end{table}

\subsection{Hand region}
\label{supp:region}

Localization by hand region is evaluated on the right hand over correctly matched predictions.
As shown in Table~\ref{tab:region}, the wrist provides the clearest separation among estimators, with PCK ranging from 0.16 to 0.57.
The differences are smaller over the fingertips and the whole hand.
The pattern is estimator-dependent: Sapiens2-1B performs substantially worse at the wrist while remaining competitive over the palm and whole hand, whereas WiLoR obtains its lowest regional score over the palm.

\subsection{Temporal stability}
\label{supp:temporal}

All temporal metrics are computed on the right hand, the side that performs the gestures; the sparse left hand carries too little continuous motion to be measured meaningfully. Three patterns stand out. 
Motion energy $E_v$ tracks jitter: the steadiest estimators also predict the least motion, so part of their stability reflects a low-motion prior rather than tracking accuracy (Table~\ref{tab:temporal_full}).
The estimator ranking is largely consistent across keypoint subsets, with the fingertips exhibiting higher jitter than the base joints, indicating that temporal instability increases toward the distal keypoints (Table~\ref{tab:temporal_subset}). 
Finally, the cleanup decreases the number of tracking gaps by approximately 37\% to 67\%, while $J_{\mathrm{jerk}}$ increases slightly in every case (Table~\ref{tab:temporal_clean}). This effect scales with the number of gaps filled: the gap-heavy estimators gain the most continuity and pay the most in jitter, while the gap-light ones barely change. The post-hoc interpolation thus trades gap continuity for a small loss of smoothness.

\begin{table}[t]
  \centering
  \footnotesize
  \setlength{\tabcolsep}{6pt}
  \renewcommand{\arraystretch}{1.15}
  \begin{tabular}{l cccc}
    \toprule
    Estimator & Gaps $\downarrow$ & $J_{\text{acc}}$ $\downarrow$ & $J_{\text{jerk}}$ $\downarrow$ & $E_v$ \\
    \midrule
    MP-holistic & 3340 & 1.159 & 1.599 & 1.524 \\
    MP-hands    & 4515 & 1.256 & 1.693 & 1.511 \\
    WiLoR       & \textbf{615}  & \textbf{0.847} & \textbf{1.181} & \textbf{1.375} \\
    RTMW-wb     & 2357 & 1.043 & 1.468 & 1.381 \\
    RTMP-hand   & 5221 & 1.312 & 1.764 & 1.427 \\
    Sapiens2-1B    & 831  & 0.937 & 1.302 & 1.399 \\
    \bottomrule
  \end{tabular}
  \caption{Temporal stability of the raw right-hand predictions over the complete videos. $J_{\text{acc}}$, $J_{\text{jerk}}$, and $E_v$ are medians over the videos scaled by 100; Gaps is the total count of gaps. $E_v$ (motion energy) contextualizes jitter and is not a quality metric.}
  \label{tab:temporal_full}
\end{table}

\begin{table}[t]
  \centering
  \footnotesize
  \setlength{\tabcolsep}{6pt}
  \renewcommand{\arraystretch}{1.15}
  \begin{tabular}{l ccc}
    \toprule
    Estimator & Whole $\downarrow$ & Base $\downarrow$ & Tips $\downarrow$ \\
    \midrule
    MP-holistic & 1.599 & 1.240 & 1.981 \\
    MP-hands    & 1.693 & 1.292 & 2.068 \\
    WiLoR       & \textbf{1.181} & \textbf{0.899} & \textbf{1.617} \\
    RTMW-wb     & 1.468 & 0.990 & 2.049 \\
    RTMP-hand   & 1.764 & 1.386 & 2.296 \\
    Sapiens2-1B    & 1.302 & 0.955 & 1.805 \\
    \bottomrule
  \end{tabular}
  \caption{$J_{\mathrm{jerk}}$ by keypoint subset for the raw right-hand predictions: whole hand, base joints (wrist and MCPs), and fingertips. The fingertips are the least stable subset for every estimator.}
  \label{tab:temporal_subset}
\end{table}

\begin{table}[t]
  \centering
  \footnotesize
  \setlength{\tabcolsep}{6pt}
  \renewcommand{\arraystretch}{1.15}
  \begin{tabular}{l cc cc}
    \toprule
    & \multicolumn{2}{c}{Gaps $\downarrow$} & \multicolumn{2}{c}{$J_{\text{jerk}}$ $\downarrow$} \\
    \cmidrule(lr){2-3}\cmidrule(lr){4-5}
    Estimator & raw & clean & raw & clean \\
    \midrule
    MP-holistic & 3340 & 1253 & 1.599 & 1.674 \\
    MP-hands    & 4515 & \textbf{1481} & 1.693 & 1.801 \\
    WiLoR       & 615  & 303  & 1.181 & 1.199 \\
    RTMW-wb     & 2357 & 966  & 1.468 & 1.517 \\
    RTMP-hand   & 5221 & \textbf{2153} & 1.764 & 1.880 \\
    Sapiens2-1B    & 831  & 524  & 1.302 & 1.309 \\
    \bottomrule
  \end{tabular}
  \caption{Effect of the full cleanup on right-hand temporal stability. The cleanup substantially reduces tracking gaps for every estimator but slightly increases $J_{\mathrm{jerk}}$. Bold marks the two largest reductions in tracking gaps.}
  \label{tab:temporal_clean}
\end{table}

\subsection{Annotation bias}
\label{supp:bias}

Table~\ref{tab:bias} quantifies the effect of evaluating MP-holistic on the bias-free subset rather than on the full annotated set.
Within the bias-free subset, manual correction moved 76\% of the keypoints by more than $\alpha{=}0.02$, with a median normalized displacement of 0.06, compared with 0.008 for biased hands.
Relative to the initialization, approximately 1{,}381 missing hands were added, and 1{,}387 spurious detections were removed from the corresponding frames.
Evaluating MP-holistic on the full set rather than the bias-free subset primarily inflates localization performance, with substantially better PCK and MPJPE.
Pose-F1 also increases, whereas Det-F1 and the label-swap rate improve only moderately.
These results show that the initialization advantage is concentrated mainly in keypoint localization, which supports reporting the main supervised comparison on the bias-free subset.

\begin{table}[t]
  \centering
  \footnotesize
  \setlength{\tabcolsep}{6pt}
  \renewcommand{\arraystretch}{1.15}
  \begin{tabular}{l ccccc}
    \toprule
    Set & Det-F1 $\uparrow$ & PCK $\uparrow$ & Pose-F1 $\uparrow$ & Swap $\downarrow$ & MPJPE $\downarrow$ \\
    \midrule
    Bias-free & 0.840 & 0.439 & 0.309 & 0.51\% & 0.265 \\
    Full set  & 0.869 & 0.579 & 0.398 & 0.35\% & 0.202 \\
    \bottomrule
  \end{tabular}
  \caption{MP-holistic evaluated on the bias-free subset and the full annotated set. Including hands that remain close to the MP-holistic initialization substantially improves PCK, MPJPE, and Pose-F1, while also improving Det-F1 and the label-swap rate. MPJPE is normalized by hand size.}
  \label{tab:bias}
\end{table}

\subsection{Benchmark size sensitivity}
\label{supp:subsampling}

To assess whether the reported comparisons depend on the size of the benchmark, we recompute the supervised metrics on nested subsets of the annotated frames. Subsampling is deterministic and applied per video: frames are grouped by video and ordered chronologically, and for a target fraction we retain $k = \mathrm{round}(n \cdot \mathrm{pct}/100)$ equispaced indices over that ordering, always including the first and last annotated frame of the clip. Sampling within each video preserves the proportion of frames across videos and subjects, and the equispaced selection keeps the subset distributed over the whole clip rather than concentrated at its beginning. 

The relative ordering of the estimators is preserved at every fraction on all four metrics, down to 10\% of the benchmark (Table~\ref{tab:subsampling}). Detection is the most stable axis, varying by less than 0.01 across all fractions, which is expected since it is computed over every annotated hand. Cross-hand overlap is equally stable: the Sapiens2 failure remains above 82\% at every fraction and stays two orders of magnitude above the other estimators, so the headline finding is not a consequence of benchmark size. Localization and Pose-F1 fluctuate more at the smallest fractions, where far fewer matched hands remain; since the subsampling is deterministic, these fluctuations reflect the particular frames retained rather than variance across repeated draws. The one effect that does depend on size is the measurement of rare failures: WiLoR's cross-hand overlap, at 0.07\% on the full benchmark, reaches exactly zero at 10\%, where too few frames remain for such infrequent events to appear at all.

\begin{table}[t]
  \centering
  \footnotesize
  \setlength{\tabcolsep}{4pt}
  \renewcommand{\arraystretch}{1.08}
  \begin{tabular}{l ccccc}
    \toprule
    Metric & 100\% & 75\% & 50\% & 25\% & 10\% \\
    \midrule
    \multicolumn{6}{l}{\textbf{Sapiens2-1B}} \\
    \quad Det-F1 $\uparrow$     & 0.666 & 0.665 & 0.665 & 0.665 & 0.666 \\
    \quad PCK $\uparrow$        & 0.491 & 0.490 & 0.481 & 0.495 & 0.488 \\
    \quad Pose-F1 $\uparrow$    & 0.361 & 0.360 & 0.359 & 0.358 & 0.360 \\
    \quad Cross-hand $\downarrow$ & 85.44\% & 85.74\% & 85.59\% & 85.78\% & 82.82\% \\
    \midrule
    \multicolumn{6}{l}{\textbf{WiLoR}} \\
    \quad Det-F1 $\uparrow$     & 0.978 & 0.978 & 0.977 & 0.975 & 0.974 \\
    \quad PCK $\uparrow$        & 0.470 & 0.475 & 0.463 & 0.454 & 0.431 \\
    \quad Pose-F1 $\uparrow$    & 0.407 & 0.409 & 0.400 & 0.414 & 0.385 \\
    \quad Cross-hand $\downarrow$ & 0.07\% & 0.10\% & 0.08\% & 0.11\% & 0\% \\
    \midrule
    \multicolumn{6}{l}{\textbf{RTMW-wb}} \\
    \quad Det-F1 $\uparrow$     & 0.932 & 0.931 & 0.934 & 0.935 & 0.926 \\
    \quad PCK $\uparrow$        & 0.560 & 0.562 & 0.541 & 0.550 & 0.581 \\
    \quad Pose-F1 $\uparrow$    & 0.477 & 0.477 & 0.460 & 0.470 & 0.496 \\
    \quad Cross-hand $\downarrow$ & 0.91\% & 0.88\% & 0.90\% & 0.76\% & 0.99\% \\
    \bottomrule
  \end{tabular}
  \caption{Sensitivity to benchmark size for three representative estimators, spanning the foundation, reconstruction-based, and whole-body paradigms. Metrics are recomputed on deterministic per-video subsets retaining the indicated fraction of annotated frames. The ordering of the estimators is preserved at every fraction on all four metrics, including the cross-hand overlap that characterizes Sapiens2.}
  \label{tab:subsampling}
\end{table}

\section{Sapiens Family Analysis}
\label{supp:sapiens}

The cross-hand overlap reported in the main paper, computed as the percentage of annotated GT hand instances affected by an overlapping opposite-handed prediction, increases sharply from Sapiens1-1B~\cite{sapiens} to Sapiens2-1B~\cite{sapiens2} (Table~\ref{tab:sapiens}).
Sapiens2-1B achieves substantially better localization and a lower label-swap rate, yet its raw Det-F1 falls well below that of Sapiens1-1B because the opposite-handed phantom predictions reduce precision.

\begin{figure}[t]
  \centering
  \includegraphics[width=\linewidth]{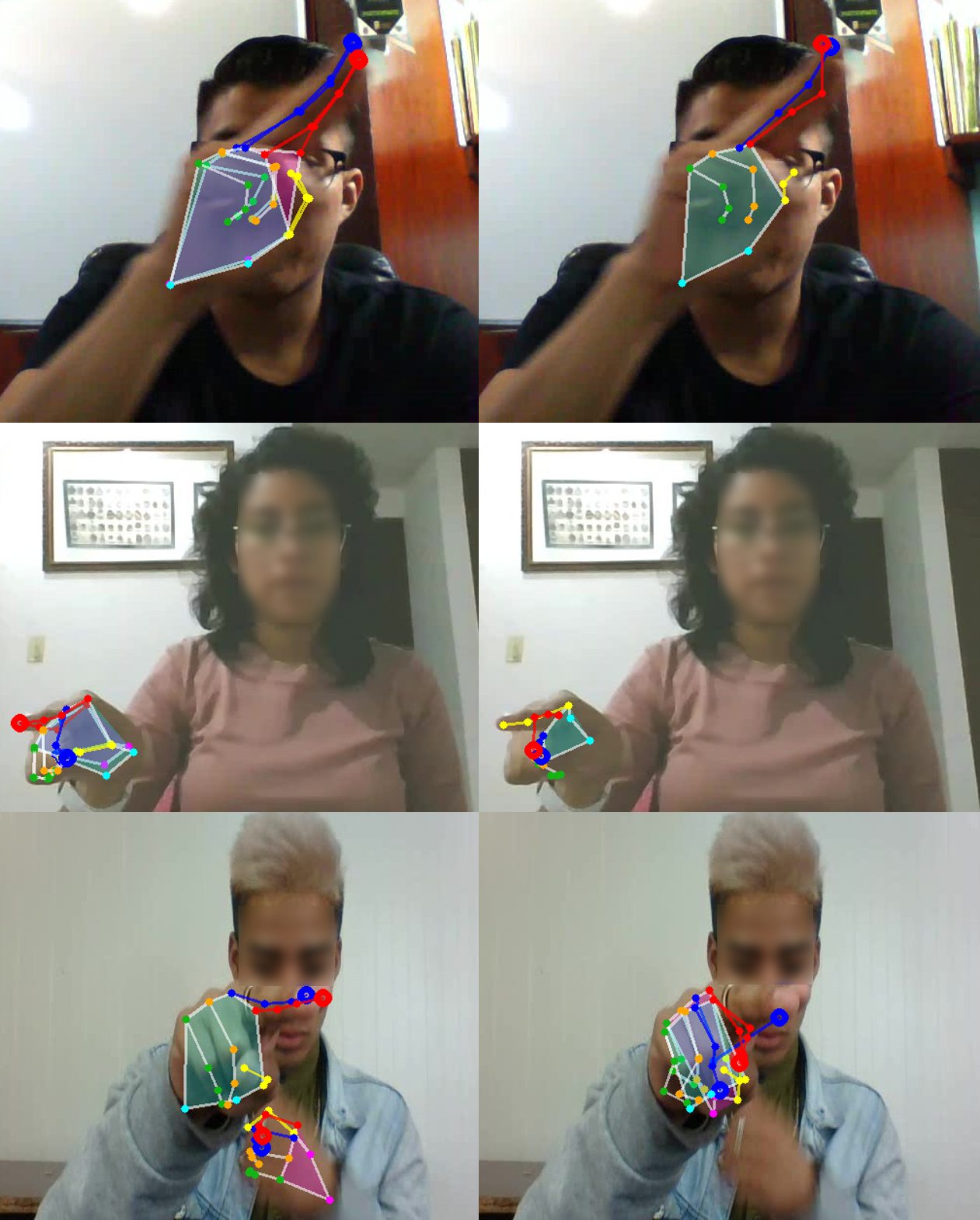}
  \caption{Qualitative comparison of Sapiens2-1B (left column) and Sapiens1-1B (right column). In the top two rows, only Sapiens2-1B produces a cross-hand overlap over the single visible hand, yet Sapiens2 localizes the keypoints markedly better.
  In the bottom row, Sapiens2-1B predicts both correctly, whereas Sapiens1-1B produces the overlap on the right hand and misses the heavily blurred left one.}
  \label{fig:sapiens_examples}
\end{figure}

Table~\ref{tab:overlap_split} groups the annotated GT hand instances according to whether their source frame contains one or both hands, while Figure~\ref{fig:sapiens_examples} shows representative predictions.
Cross-hand overlap is rare when both hands are present, but becomes dominant for Sapiens2-1B in single-hand frames.
This pattern indicates that the failure primarily consists of an opposite-handed duplicate prediction over the visible hand rather than a failure to recover the visible hand itself.
The behavior may reflect a full-body prediction prior that favors bilateral hand outputs even when only one hand is visible.
Because VERA-Hand consists predominantly of single-hand frames, it provides a setting in which this failure becomes directly measurable.

The cleanup results further isolate the source of the performance loss.
The spatial stage removes the cross-hand overlap entirely and raises Det-F1 from 0.666 to 0.929, while PCK and MPJPE remain effectively unchanged.
The temporal stage provides a smaller additional gain in detection and removes the remaining duplicate-hand predictions.
These results indicate that the dominant Sapiens2-1B failure in this benchmark is spurious hand recovery rather than inaccurate localization of the visible hand.

\begin{table*}[t]
  \centering
  \footnotesize
  \setlength{\tabcolsep}{6pt}
  \renewcommand{\arraystretch}{1.2}
  \begin{tabular}{l ccccccc}
    \toprule
    Variant & Cross-hand $\downarrow$ & Dup. $\downarrow$ & Det-F1 $\uparrow$ & PCK $\uparrow$ & Pose-F1 $\uparrow$ & Swap $\downarrow$ & MPJPE $\downarrow$ \\
    \midrule
    Sapiens1-1B                       & 14.15\% & 2.95\% & 0.837 	& 0.360 	& 0.221 & 0.51\% & 0.230 \\
    Sapiens2-1B                       & 85.44\% & 1.85\% & 0.666 	& 0.491 	& 0.361 & 0.03\% & 0.172 \\
    \quad + spatial                   & 0\%     & 0.98\% & 0.929 	& \textbf{0.492} & 0.402 & 0.09\% & \textbf{0.171} \\
    \quad + spatial + temporal        & 0.05\%  & 0\%    & \textbf{0.940} & 0.490 	& \textbf{0.406} & 0.03\% & 0.175 \\
    \bottomrule
  \end{tabular}
  \caption{Comparison of the Sapiens family and the effect of applying the cleanup to Sapiens2-1B. The cross-hand overlap worsens from Sapiens1 to Sapiens2, and the two-stage cleanup applied to Sapiens2 (spatial, then spatial plus temporal) removes it and recovers detection performance while leaving PCK and MPJPE nearly unchanged.}
  \label{tab:sapiens}
\end{table*}

\begin{table}[t]
  \centering
  \footnotesize
  \setlength{\tabcolsep}{6pt}
  \renewcommand{\arraystretch}{1.2}
  \begin{tabular}{l c cc}
    \toprule
    Frame context & GT hands & Sapiens1-1B & Sapiens2-1B \\
    \midrule
    Both hands present   & 366     & 2.25\%  & 2.25\% \\
    Single hand present  & 10{,}324 & 14.56\% & \textbf{88.35\%} \\
    All annotated hands  & 10{,}690 & 14.15\% & 85.44\% \\
    \bottomrule
  \end{tabular}
  \caption{Cross-hand overlap grouped by whether the source frame contains one or both hands. Overlap is rare when both hands are present, but it explodes in single-hand frames for Sapiens2-1B.}
  \label{tab:overlap_split}
\end{table}

\section{Post-hoc Cleanup: Method and Ablation}
\label{supp:cleanup}

The cleanup operates exclusively on the predicted hands and uses no GT, image content, or localization metric.
It comprises a spatial stage that resolves overlapping opposite-handed predictions within individual frames and a temporal stage that enforces short-term consistency across consecutive frames.

\textbf{Spatial stage.} For every pair of opposite-handedness predictions in a frame, we compute the bounding-box IoU (from the valid keypoints) and the center distance normalized by the larger box diagonal. The pair is treated as a cross-hand overlap when IoU $\geq 0.80$ and the normalized center distance $\leq 0.15$, in which case the hand with the higher mean keypoint confidence is kept and the other removed. 
The spatial stage primarily addresses the cross-hand overlap problem.

\textbf{Temporal stage.} The per-video timeline is independently processed by hand over a sliding odd window of $7$ frames (a radius of $3$) using a binary presence signal, with three components. 
Deduplication retains at most one prediction per hand label in each frame, selecting the prediction with the most valid keypoints.
Phantom removal discards a prediction supported by fewer than $2$ neighbors in its window, treating it as a non-persistent detection. 
Gap recovery fills an absent side that has at least $4$ supporting neighbors, with support on both temporal directions, by linearly interpolating the keypoints between the nearest present frames; it is restricted to gaps of at most $3$ frames and skipped when the motion between the endpoints is too large, to avoid interpolating across fast movement. 
The complete cleanup applies the spatial stage first, followed by all three components of the temporal stage.

\textbf{Ablation.} We ablate the temporal stage on MP-hands, whose gain is purely temporal since it has no cross-hand overlap, and compare against a majority-vote baseline that forces each side present whenever it appears in more than half of the window, copying the nearest hand without interpolation.
Table~\ref{tab:ablation} shows that deduplication removes the duplicate predictions but produces little change in the evaluation scores. Gap recovery provides the main individual gain in detection, although it also reduces PCK by introducing interpolated predictions, whereas phantom removal improves PCK and Pose-F1 at a small cost in detection by discarding non-persistent detections.
Combining the three components outperforms every individual variant on both Det-F1 and Pose-F1, indicating that the components are complementary and that the detection gains outweigh the accompanying localization change.
The majority-vote baseline, which forces presence without interpolation, remains below the full temporal stage on all three metrics.

\begin{table}[t]
  \centering
  \footnotesize
  \setlength{\tabcolsep}{5pt}
  \renewcommand{\arraystretch}{1.2}
  \begin{tabular}{l cccc}
    \toprule
    Variant & Det-F1 $\uparrow$ & PCK $\uparrow$ & Pose-F1 $\uparrow$ & Dup. $\downarrow$ \\
    \midrule
    Raw                         & 0.898 & 0.573 & 0.389 & 1.48\% \\
    \quad + Deduplication       & 0.899 & 0.570 & 0.390 & 0\% \\
    \quad + Phantom removal     & 0.879 & \textbf{0.580} & 0.430 & 1.47\% \\
    \quad + Gap recovery        & 0.913 & 0.556 & 0.400 & 1.48\% \\
    \quad + Full temporal stage & \textbf{0.926} & 0.559 & \textbf{0.442} & 0\% \\
    \midrule
    Majority baseline           & 0.917 & 0.537 & 0.436 & 1.46\% \\
    \bottomrule
  \end{tabular}
  \caption{Temporal-stage ablation on MP-hands using the bias-free subset. Individual components are applied independently to the raw predictions, while the full temporal stage applies all three. The full temporal stage achieves the highest Det-F1 and Pose-F1, while phantom removal yields the highest PCK.}
  \label{tab:ablation}
\end{table}

Together, these stages produce the cleaned results reported throughout the paper.
The spatial stage drives the Sapiens2-1B recovery shown in Table~\ref{tab:sapiens}, while the temporal stage produces the reduction in tracking gaps and the accompanying jitter increase reported in Table~\ref{tab:temporal_clean}.
Applying both stages yields the raw-versus-clean comparison presented in Table~\ref{tab:cleanup}.

\section{Qualitative Results}
\label{supp:qualitative}

\begin{figure*}[h]
  \centering
  \includegraphics[width=\textwidth]{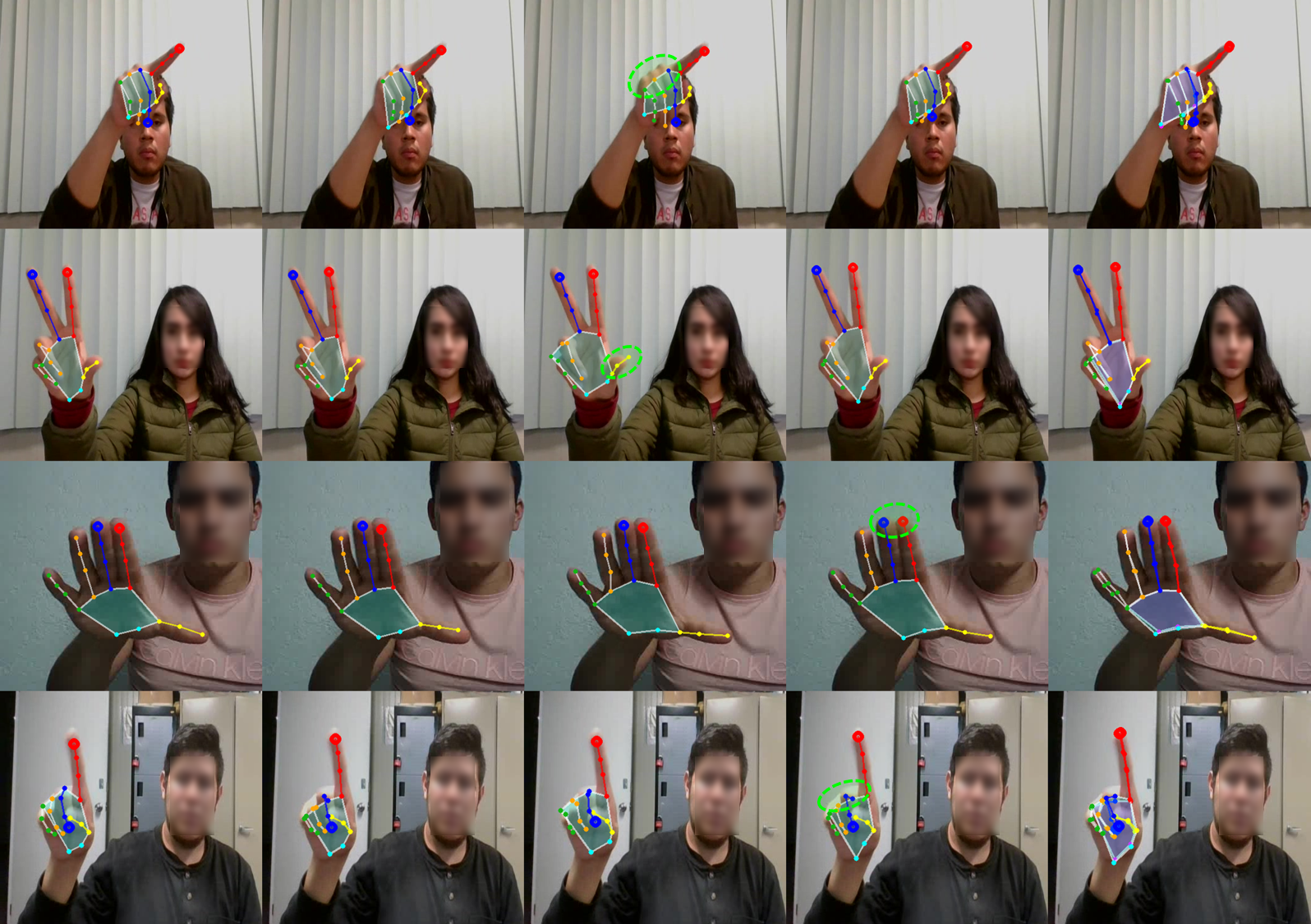}
  \caption{Qualitative comparison on frames where the estimators largely agree with the GT. Columns, from left to right: GT, MP-hands, WiLoR, RTMW-wb, and Sapiens2-1B. Every hand is recovered by every pipeline, and green dashed ellipses mark the residual localization deviations, which remain confined to individual keypoints. 
  }
  \label{fig:qual_ok}
\end{figure*}

\begin{figure*}[h]
  \centering
  \includegraphics[width=\textwidth]{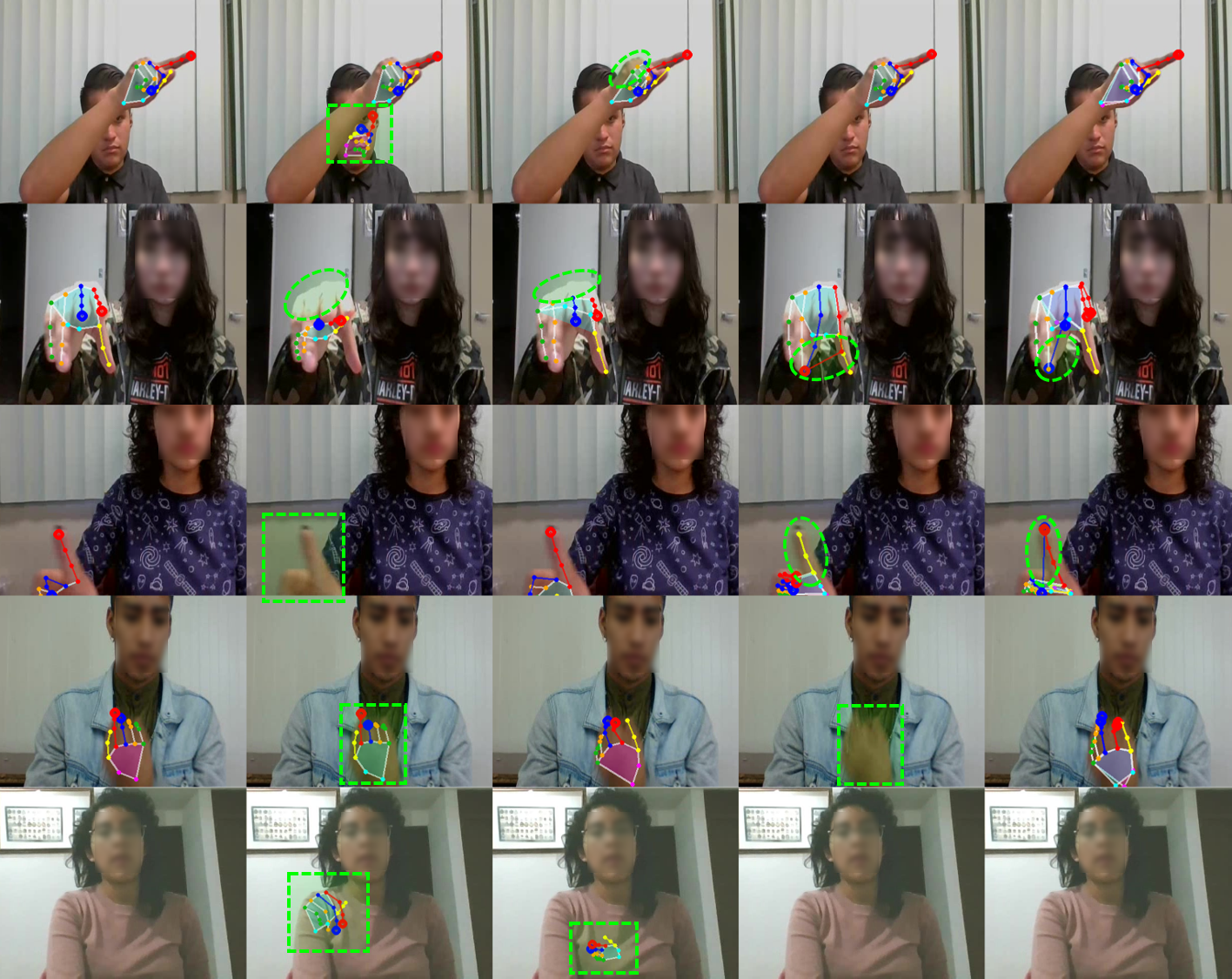}
  \caption{Qualitative comparison on frames where one or more estimators fail. Columns, from left to right: GT, MP-hands, WiLoR, RTMW-wb, and Sapiens2-1B. Green dashed rectangles mark detection or identification errors, and green dashed ellipses mark gross localization errors on a recovered hand. The last row shows a frame without hands in which two pipelines still emit a prediction. Failures are pipeline-dependent: a frame that defeats one estimator is often handled correctly by another, and the errors shown here are invisible to an evaluation restricted to matched predictions.}
  \label{fig:qual_no}
\end{figure*}

This section complements the quantitative analysis with side-by-side comparisons between the ground truth and the predictions of the four representative estimators shown in Figure~\ref{fig:blur}. Each row shows the same frame under all five conditions, in the order GT, MP-hands, WiLoR, RTMW-wb, and Sapiens2-1B. Figure~\ref{fig:qual_ok} shows frames on which the estimators largely agree with the GT, while Figure~\ref{fig:qual_no} shows frames on which one or more pipelines fail.

The two figures illustrate the separation between the evaluation axes. In Figure~\ref{fig:qual_ok}, every hand is recovered by every pipeline and the remaining discrepancies are confined to individual keypoints, typically at the fingertips or the wrist, which is the regime that localization metrics such as PCK are designed to capture and the only one visible to a localization-only comparison. Figure~\ref{fig:qual_no} shows the failure modes that such a comparison cannot express: hands that are missed entirely, spurious predictions placed on the forearm or the torso, and gross pose configurations that are wrong even though a hand was found at approximately the right location. The last row is an example of a no-hand frame, in which two pipelines still emit a hand; these frames contribute no localization measurement at all and are only scored by the detection axis.

The examples also show that failures are not evenly distributed across pipelines. A frame that is difficult for one estimator is frequently handled correctly by another, and no column is consistently the weakest across rows, which is the qualitative counterpart of the ranking reversals reported in Section~\ref{sec:results}. Motion blur, hands overlapping the face, and hands that partially leave the frame account for most of the cases in Figure~\ref{fig:qual_no}, consistent with the stratified results in Section~\ref{supp:results}.

\end{document}